\documentclass[runningheads]{llncs}
\usepackage[T1]{fontenc}
\usepackage{graphicx}
\usepackage[colorlinks=true,linkcolor=blue,citecolor=blue,urlcolor=blue]{hyperref}
\usepackage{color}

\begin{document}
\title{%TimeDiff for Irregular Clinical Labs: Joint Generation of Values and Informative Missingness on DACMI (MIMIC-III)
% Modeling Informative Missingness in Irregular Clinical Labs with Diffusion Models
Informative Missingness to Generate Irregular Clinical Time Series
}
\author{Hadi Mehdizavareh\inst{1}\thanks{Corresponding author.} \and
Gabriele Santangelo\inst{2} \and
Giovanna Nicora\inst{2} \and
Simon Lebech Cichosz\inst{1} \and
Arianna Dagliati\inst{2} \and
Arijit Khan\inst{3, 1} \and
Riccardo Bellazzi\inst{2,1}}
\authorrunning{M. H. Mehdizavareh et al.}
\institute{
Aalborg University, Aalborg, Denmark\\
\email{mhme@cs.aau.dk}
% \email{mhme@cs.aau.dk, simcich@hst.aau.dk, arijitk@bgsu.edu}
\and
University of Pavia, Pavia, Italy\\
% \email{gabriele.santangelo01@universitadipavia.it, giovanna.nicora@unipv.it, arianna.dagliati@unipv.it, riccardo.bellazzi@unipv.it}
\and
Bowling Green State University, Bowling Green, Ohio, USA\\
% \email{arijitk@bgsu.edu}
}
\maketitle              % typeset the header of the contribution
\begin{abstract}
%Laboratory tests in electronic health records are taken irregularly, and when a test is not ordered can be as informative as the result itself. Missingness therefore, reflects clinical decisions and patient physiology, and should be modeled rather than treated as a preprocessing artifact. We revisit diffusion-based time-series generation in this regime by jointly modeling lab values and informative missingness on the public DACMI benchmark derived from MIMIC-III. To preserve realistic sampling behavior while enabling tractable diffusion training, we regularize chart times into 4-hour bins, limiting artificial missingness introduced by discretization, and segment admissions into non-overlapping 7-day windows. Each window is represented as a 26-channel trajectory comprising 13 continuous lab variables and 13 binary observation indicators, allowing the model to learn a sequence-level joint distribution over values and missingness. We further apply monotonic transformations (log or Box--Cox) and z-score normalization to move lab marginals toward Gaussianity. Building on TimeDiff, we modify the Gaussian loss to exclude truly missing numerical entries so gradients are driven only by observed measurements, while the missingness process is learned directly via the discrete diffusion objective. Experiments show close alignment between real and synthetic data in both univariate lab distributions and joint embeddings over values plus masks, indicating that diffusion models can capture coupled physiology and sampling patterns under MNAR-like missingness.

Laboratory tests in electronic health records are collected irregularly, and the absence of a test order can be as informative as the measurement itself. Such missingness reflects clinicians' decisions and patient physiology, making it important to model it directly rather than treat it as a preprocessing artifact. Here we present a diffusion‑based approach for generating clinical time series that jointly models laboratory values and their observation patterns using the public Data Analytics Challenge on Missing data Imputation (DACMI) benchmark derived from MIMIC‑III. To preserve realistic sampling, we align chart times into 4‑hour intervals and segment admissions into 7‑day windows, producing trajectories that pair each lab value with a corresponding observation indicator. Standard transformations and normalization are applied to stabilize training. Our method extends the TimeDiff framework to learn continuous lab values and discrete missingness patterns through complementary diffusion objectives. Experiments show that the generated data closely match real patient trajectories across individual lab distributions and joint value-missingness embeddings, demonstrating that diffusion models can capture clinically meaningful dependencies between patient physiology and clinicians’ testing behavior under MNAR‑like (missing‑not‑at‑random) missingness. These preliminary results indicate that our model can serve as an initial component toward developing clinical foundation models. By producing synthetic priors that preserve key physiology--missingness relationships, this work motivates the subsequent training of Prior-Data Fitted Networks capable of leveraging informative missingness, which we will investigate in the extended work.

\end{abstract}

\section{Background}
Electronic health records (EHRs) record irregular sequences of measurements shaped by physiology, patient behavior, and clinical decisions \cite{xiao_opportunities_2018}. As an example, labs are rarely collected on a fixed schedule, and the absence of testing can itself reflect severity or clinical intent. Prior work shows that lab missingness is often informative rather than random and can differ across sites and care settings \cite{tan_informative_2023,cichosz_biases_2026}. For generative modeling, this means that producing realistic synthetic lab trajectories requires matching not only lab values, but also when tests occur, how they appear in bursts, and how sampling relates to the underlying values.
Many clinical time-series pipelines handle missingness mainly through imputation, aiming to densify sparse records before modeling \cite{kazijevs_deep_2023,li_imputation_2021}. The DACMI challenge \cite{luo_evaluating_2022}, built from MIMIC-III \cite{johnson_mimic-iii_2016}, offers a public benchmark for imputation methods with ground truth for 13 common blood labs. However, success on imputation benchmarks does not necessarily show that a generator captures the joint process of values and observations under real EHR sampling. This is most critical in the Missing-Not-At-Random (MNAR) setting, where whether a lab is measured depends on unobserved or partially observed clinical state, making missingness tightly coupled to the data-generating mechanism.
Diffusion models are state-of-the-art option for time-series generation \cite{ho_denoising_2020,yang_survey_2026}. TimeDiff \cite{tian_reliable_nodate} is among the first diffusion approaches tailored to EHR sequences and supports mixed continuous and binary variables. Still, much prior evaluation focuses on near-regular signals (e.g., vitals) and reports missingness mainly through marginal rates, leaving open whether models reproduce irregular lab sampling bursts and their coupling with values—an issue that can persist even when univariate distributions look correct.
\newline\noindent\textbf{Contributions.} We present a generative framework designed as the preliminary step for developing clinical foundation models, providing high-fidelity synthetic priors required to train Prior-Data Fitted Networks \cite{muller_transformers_2021}. Our contributions are: \textbf{(i)} A diffusion-based synthetic lab generation pipeline on a public benchmark that targets irregular labs and informative missingness, generating both lab values and their observation indicators jointly to support downstream inference. \textbf{(ii)} We modify the numerical (Gaussian) loss in TimeDiff \cite{tian_reliable_nodate} to exclude truly missing lab values, treating missingness as absence rather than noisy values. \textbf{(iii)} We present a practical preprocessing and windowing protocol designed to preserve MNAR-like structure while remaining compatible with diffusion training. \textbf{(iv)} We evaluate distributional similarity using univariate lab distributions, joint embeddings over values and masks, and additional similarity metrics for model selection to assess realism in both physiology and sampling behavior.
\section{Materials and Methods}
\textbf{Dataset.} We used the publicly available DACMI imputation challenge dataset \cite{luo_evaluating_2022}, derived from MIMIC-III, which provides a standardized train/test split designed for fair comparison of methods on sparse clinical time series. From each admission, we extracted 13 laboratory variables and constructed a corresponding missingness mask directly from the raw tables, where each lab at each time step is marked as observed (0) or missing (1). This yields a paired representation of physiology (continuous lab values) and sampling behavior (binary indicators) that is suitable for studying informative missingness, especially in the MNAR-like regime where ordering and frequency of tests reflect clinical decisions. Models development and selection are based on the DACMI training split.
\newline\noindent\textbf{Setup and preprocessing.} We discretized \texttt{charttime} (minutes since admission) into 4-hour intervals to minimize artificial missingness, creating 42-step windows (7 days). Admissions exceeding this horizon were segmented into consecutive non-overlapping windows, yielding 16,580 training samples from 8,267 admissions. To approximate Gaussianity, features underwent log or Box-Cox transformations followed by z-score standardization. The model input consists of 26 channels (13 standardized values and 13 binary masks). In addition to the values and binary mask trajectories, we used \emph{delta conditioning} to encode observation recency. The delta conditioning provides an auxiliary conditioning tensor to the denoising network at each diffusion step. Therefore, the model still generates only the lab values and mask variables, but is guided by this extra context during denoising. For each lab and time index, the delta value represents the number of steps since the most recent observation. Specifically, at each time index, we computed the number of time steps that must be traversed backward from the current position to reach the latest time point at which the lab is observed (mask $=0$). This yields a $(42, 13)$ matrix per window, normalized to $[0,1)$ by division by 42.
\newline\noindent\textbf{Experimental setup and sampling.} We trained the model for 200,000 steps using the Adam optimizer (learning rate $8\times10^{-5}$, batch size 64). To balance our two objectives, we applied a weight of $\lambda=10^{-5}$ to the discrete missingness loss relative to the continuous value loss, adopting the notation from TimeDiff~\cite{tian_reliable_nodate}. For evaluation, we generated 16,000 synthetic windows (42 steps $\times$ 26 channels) by iteratively denoising random noise. This sample size matches the real dataset, allowing us to compare the synthetic distributions and observation patterns against the ground truth.
\section{Results and Discussions}
\begin{figure}[!t]
    \centering
    \includegraphics[width=\textwidth]{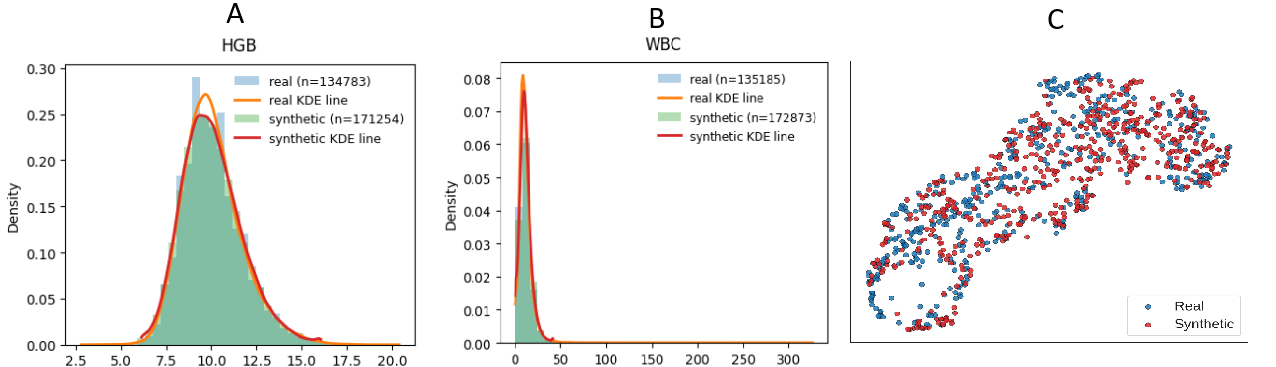}
    \caption{\textbf{Real vs.\ synthetic comparisons:} (A) Hemoglobin and (B) White Blood Cells count univariate densities; (C) UMAP embedding of windows using all 26 channels (labs + masks), computed on a random subsample of 500 real and 500 synthetic windows.}
    \label{fig:univariate_umap}
\end{figure}

Figure \ref{fig:univariate_umap}, panels A--B compare real and synthetic univariate distributions for two representative labs with different shapes. Hemoglobin (A) is fairly concentrated with a mild right tail, and the synthetic density closely matches the real histogram around the main mode and across the upper tail. White Blood Cells count (B) is markedly more right-skewed with a long tail driven by occasional high values; despite this more challenging setting, the synthetic samples follow the real distribution well in the high-density region and preserve the overall tail profile. 
Figure \ref{fig:univariate_umap} panel C compares the overall distribution of real data with the synthetic ones. The embedding was computed using UMAP with a DTW-based distance metric, for all the 26 channels (13 values and 13 masks). The substantial overlap between the two distributions indicates that the generative approach produces high-quality synthetic samples, without generating instances outside the support of the real data.
\begin{table}[!tbp]
\centering
\caption{Per-feature comparison across 13 labs values: missingness prevalence (mask$=1$), univariate EMD on lab-value marginals, and DTW trajectory distance.}
% \caption{Per-feature comparison of real (R) and synthetic (S) trajectories. Missingness prevalence is the fraction of mask$=1$ entries. EMD compares univariate lab-value distributions, and DTW compares feature-level trajectories on $N=100$ windows.}
\label{tab:combined_metrics}
\scriptsize
\setlength{\tabcolsep}{3pt}
\renewcommand{\arraystretch}{1.05}
\begin{tabular}{p{3.6cm}ccccccccc}
\hline
& \multicolumn{3}{c}{Missingness prevalence} & \multicolumn{3}{c}{EMD (univariate)} & \multicolumn{3}{c}{DTW (trajectory)} \\
\cline{2-4}\cline{5-7}\cline{8-10}
Lab
& Real
& \shortstack{Synth\\w/ $\Delta$}
& \shortstack{Synth\\w/o $\Delta$}
& R--R
& \shortstack{R--S\\w/ $\Delta$}
& \shortstack{R--S\\w/o $\Delta$}
& R--R
& \shortstack{R--S\\w/ $\Delta$}
& \shortstack{R--S\\w/o $\Delta$} \\
\hline
Chloride   & 0.78 & 0.72 & 0.74 & 0.06 & 0.09 & 0.11 & 37.71 & 37.63 & 38.09 \\
Potassium    & 0.78 & 0.72 & 0.74 & 0.03 & 0.06 & 0.09 & 38.44 & 38.27 & 38.48 \\
Bicarb & 0.78 & 0.72 & 0.74 & 0.07 & 0.14 & 0.14 & 37.90 & 37.80 & 38.13 \\
Sodium   & 0.78 & 0.72 & 0.74 & 0.05 & 0.12 & 0.09 & 37.91 & 37.77 & 38.20 \\
Hematocrit   & 0.80 & 0.75 & 0.77 & 0.06 & 0.06 & 0.03 & 38.34 & 38.83 & 38.82 \\
Hemoglobin   & 0.81 & 0.76 & 0.78 & 0.02 & 0.11 & 0.07 & 38.42 & 38.97 & 39.00 \\
MCV   & 0.81 & 0.76 & 0.78 & 0.03 & 0.16 & 0.20 & 37.73 & 38.00 & 38.12 \\
Platelets   & 0.81 & 0.76 & 0.77 & 0.08 & 0.20 & 0.24 & 37.72 & 37.62 & 37.83 \\
WBC count   & 0.81 & 0.76 & 0.77 & 0.08 & 0.07 & 0.08 & 37.88 & 38.01 & 38.15 \\
RDW   & 0.81 & 0.76 & 0.78 & 0.04 & 0.09 & 0.13 & 37.53 & 37.53 & 37.88 \\
BUN  & 0.78 & 0.72 & 0.74 & 0.08 & 0.10 & 0.15 & 36.41 & 36.70 & 37.10 \\
Creatinine  & 0.78 & 0.72 & 0.74 & 0.02 & 0.06 & 0.08 & 35.99 & 35.87 & 36.40 \\
Glucose  & 0.78 & 0.72 & 0.74 & 0.03 & 0.07 & 0.04 & 37.51 & 37.43 & 37.72 \\
\hline
Mean  & 0.79 & 0.74 & 0.76 & 0.05 & 0.10 & 0.11 & 37.65 & 37.72 & 37.99 \\
\hline
\end{tabular}
\end{table}

Table~\ref{tab:combined_metrics} summarizes per-feature realism across the 13 lab variables. It reports (i) missingness prevalence (fraction of time points with mask$=1$) in real windows and in synthetic windows generated with and without delta conditioning, (ii) univariate Earth Mover's Distance (EMD) \cite{farahani_time-series_2025} between lab-value marginals using Real--Real as a same-domain reference and Real--Synthetic for each generator, and (iii) Dynamic Time Warping (DTW) trajectory distances computed per feature on a random subsample of $N=100$ windows. Both EMD and DTW are computed in a missingness-aware manner by restricting comparisons to time points marked as observed according to the real or generated masks.

Overall, delta conditioning yields slightly better agreement with the real data: the mean EMD and DTW are lower than the corresponding no-conditioning setting and closer to the Real--Real reference. In contrast, the synthetic missingness prevalence is consistently below the observed prevalence, suggesting that the model still underestimates the frequency of unobserved measurements and that the missingness generation process could be improved. Finally, we note that these results should be interpreted as preliminary evidence that the model captures important aspects of the coupled physiology--sampling process, rather than as a complete validation of downstream clinical utility. In the extended work, we evaluate how the resulting synthetic data can be used to train Prior-Data Fitted Networks for imputation and forecasting under distribution shifts, while explicitly leveraging informative missingness.

\vspace{-2mm}
\section{Conclusions}
This preliminary work shows how TimeDiff can be adapted to generate laboratory trajectories while jointly modeling informative missingness on a public MIMIC-III–derived benchmark. With practical preprocessing, the model produces synthetic windows that closely match real data in marginal lab distributions and in the joint structure of values and missingness masks. Our results suggest that diffusion models are a promising foundation for realistic synthetic lab generation under MNAR-like missingness, and they motivate future work that leverages the learned generative prior for downstream inference and forecasting in foundation models.

\section{Acknowledgment}
Hadi Mehdizavareh and Arijit Khan acknowledge support from the Novo Nordisk Foundation grant NNF 22OC0072415.

\vspace{-2mm}

\end{document}